\documentclass[runningheads,a4paper]{llncs}

\usepackage{hyperref}
\usepackage{amssymb}
\usepackage{graphicx}
\usepackage{subfloat}
\usepackage{subfig}
\usepackage[font=small,skip=5pt]{caption}
\usepackage{color}
\usepackage{wrapfig,booktabs}
\usepackage{adjustbox}

\usepackage{url}
\urldef{\mailsa}\path|saad.akram@oulu.fi|    
\newcommand{\keywords}[1]{\par\addvspace\baselineskip
\noindent\keywordname\enspace\ignorespaces#1}

\begin{document}

\mainmatter  % start of an individual contribution

\title{Leveraging Unlabeled Whole-Slide-Images for Mitosis Detection}
\titlerunning{Leveraging Unlabeled WSIs for Mitosis Detection}
% LUMPE: Leverage Unlabeled Mitotic Cells for Performance Enhancement

%\iffalse
\iftrue
\author{Saad Ullah Akram\textsuperscript{1,2}, Talha Qaiser\textsuperscript{2}, Simon Graham\textsuperscript{2}, Juho Kannala\textsuperscript{3}, Janne Heikkil\"{a}\textsuperscript{1}, and Nasir Rajpoot\textsuperscript{2,4}%
}
\authorrunning{Akram, et.al.}

\institute{\textsuperscript{1}Center for Machine Vision and Signal Analysis (CMVS),\\
	University of Oulu, Finland\\
	\textsuperscript{2}Tissue Image Analytics (TIA) Lab, University of Warwick, UK\\
	\textsuperscript{3}Department of Computer Science, Aalto University, Finland\\
	\textsuperscript{4}The Alan Turing Institute, UK\\
	\mailsa}

\else 
\iffalse
\author{xxxx\textsuperscript{1}%
\institute{\textsuperscript{1}yyyy\\}
}
\fi
\fi

%\toctitle{Lecture Notes in Computer Science}
%\tocauthor{Authors' Instructions}
\maketitle

% 70-0.550 words
\begin{abstract}
Mitosis count is an important biomarker for prognosis of various cancers.
At present, pathologists typically perform manual counting on a few selected regions of interest in breast whole-slide-images (WSIs) of patient biopsies.
This task is very time-consuming, tedious and subjective.
Automated mitosis detection methods have made great advances in recent years.
However, these methods require exhaustive labeling of a large number of selected regions of interest.
This task is very expensive because expert pathologists are needed for reliable and accurate annotations.
In this paper, we present a semi-supervised mitosis detection method which is designed to leverage a large number of unlabeled breast cancer WSIs.
As a result, our method capitalizes on the growing number of digitized histology images, without relying on exhaustive annotations, subsequently improving mitosis detection. 
Our method first learns a mitosis detector from labeled data, uses this detector to mine additional mitosis samples from unlabeled WSIs, and then trains the final model using this larger and diverse set of mitosis samples.
The use of unlabeled data improves F1-score by $\sim$5\% compared to our best performing fully-supervised model on the TUPAC validation set.
Our submission (single model) to TUPAC challenge ranks highly on the leaderboard with an F1-score of 0.64.
\end{abstract}

\setlength{\textfloatsep}{1.2mm}
\section{Introduction}
\vspace{-0.3em}
Precise quantification of mitotic figures in hematoxylin and eosin (H\&E) stained slides is of high clinical significance in understanding the proliferation activity of cells within tumor regions.
In breast cancer, mitosis count is a significant prognostic biomarker and one of the most important criteria for cancer grading.
In routine clinical practice, a pathologist visually examines H\&E stained slides under the microscope.
This conventional way of mitosis counting is extremely time-consuming and tedious, as a pathologist has to perform this task for several high-power-fields (HPF) in multiple whole-slide-images (WSI) for a single patient.
In addition, this process is extremely subjective with noticeable disagreement between different pathologists \cite{Veta2016}, mainly due to the inherent ambiguity and difficulty of the task.
Automated methods have the potential to overcome the inter- and intra-observer variability by producing more consistent results and freeing up valuable pathologist's time which can be better spent on understanding the aggressiveness of disease and stratified medicine.

Mitosis detection is considered a challenging task even for experienced pathologists due to the variations in morphological appearance of mitotic cells (see Fig.~\ref{fig:datasets}).
These variations are caused by various factors including the mitotic phase, staining variability and tissue damage during the slide preparation. 
Mitotic cells typically look like hyper-chromatic objects without nuclear membrane, which have hairy extensions of nuclear material \cite{Veta2016}.
However, there are many instances in which these characteristic features are difficult to spot and subjective decisions have to be made.
These subjective decisions can be challenging due to the presence of other cells, such as apoptotic (programmed cell death) cells, which have very similar appearances.
In addition, mitotic cells are significantly less in number as compared to other malignant and healthy cells.
Hence, the detection of mitotic cells naturally suffers from a huge class imbalance.

In recent years, several mitosis detection challenges have been organized: MITOS12 \cite{Roux2013}; AMIDA13 \cite{Veta2015}; MITOS14 \cite{mitos}; and TUPAC \cite{tupac}.
These challenges have tackled increasingly more difficult scenarios and have greatly helped to advance mitosis detection research.
Regardless, one of the shortcomings of these contest datasets is the limited number of mitosis samples and pathology centers from which the data is acquired, limiting the generalization ability of trained mitosis detectors.
State-of-the-art deep learning based methods require a large number of annotated samples for training.
One approach for increasing the number of annotations is semi-supervised learning, which can make good use of readily available unlabeled data in histopathology.

With regards to the previously mentioned challenges, we propose a deep learning based self-supervised algorithm that makes use of both labeled HPFs and unlabeled WSIs to train the mitosis detector.
We show that mitosis detection can benefit from larger datasets and in the absence of a large amount of labeled data, leveraging an unlabeled dataset can lead to a significant improvement in the performance of current mitosis detection methods.

\vspace{-0.6em}
\section{Related research}
\vspace{-0.6em}
Automated methods have shown great promise in detecting mitotic cells \cite{Ciresan2013,Khan2013,Paeng2017} within histological images in recent years.
Mitosis detection was among the first few problems where deep learning showed its potential, when a convolutional neural network (CNN) \cite{Ciresan2013} out-performed competing methods by a large margin in MITOS12 challenge \cite{Roux2013}.
This method detected mitosis by scanning an image in a sliding window manner.
This was followed by a fully convolutional version, which significantly reduced the test inference time and won the AMIDA13 challenge by a convincing margin \cite{Veta2015}.

Since then, there have been many advances to improve the performance of these methods.
Some of the most recent methods have made use of deeper \cite{Paeng2017}, wider \cite{Zerhouni2017} and cascades of networks \cite{Chen2016}.
%Object detection based methods such as faster R-CNN \cite{Li2018} have also been applied to this task with some success.
Recently, the use of crowd-sourced annotations \cite{Albarqouni2016} has been explored for this task and despite these annotations being noisy, the proposed aggregation layer can lead to good performance.
PHH3 labeling has also been used to create larger mitosis detection datasets \cite{Tellez2018}.

In the last couple of years, semi- and weakly-supervised methods have shown their potential in improving performance beyond the limits of state-of-the-art fully-supervised methods.
Sun et al. \cite{Sun2017} showed the importance of dataset size and made use of noisy web-scale dataset in addition to ImageNet to achieve state-of-the-art performance on various important vision tasks including image classification, object detection and segmentation.
Radosavovic et al. \cite{Radosavovic2017} made use of self-supervision and a large unlabeled dataset to improve performance on object and human keypoint detection tasks.
These methods are able to utilize large unannotated datasets, in addition to annotated datasets, and improve performance over fully-supervised techniques.

\vspace{-0.6em}
\section{Method}
\vspace{-0.6em}
Our method consists of 3 stages shown in Fig.~\ref{fig:nets}. 
First, we train a mitosis detector on exhaustively labeled high-power-fields (HPFs) (Section \ref{method:detector});
second, we mine unannotated whole-slide-images (WSIs) for additional mitosis patches (Section \ref{method:mining}); and finally, we re-train the mitosis detector using patches from both HPFs and WSIs (Section \ref{method:detector}).

\vspace{-0.6em}
\subsection{Mitosis detector}
\vspace{-0.4em}
\label{method:detector}
We use a 12-layer ResNet \cite{He2015a}, shown in Fig.~\ref{fig:nets}b, for mitosis detection.
\vspace{-0.8em}
\subsubsection{Training:}
\begin{figure*}[!tb]
	\centering
	\includegraphics[width=12.0cm]{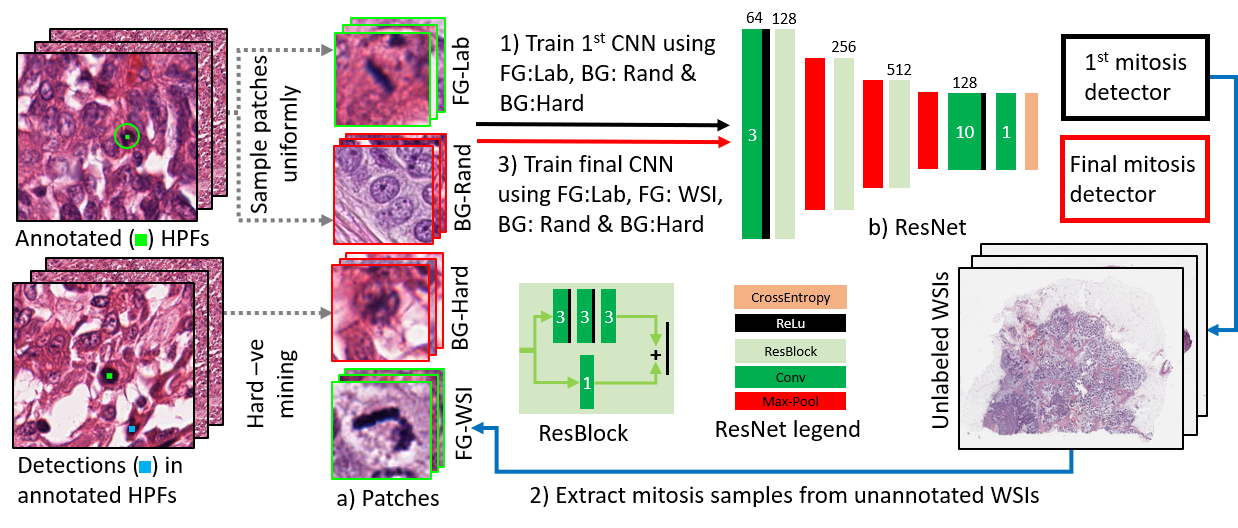}
	\vspace{-0.2em}
	\caption{Overview of the method.
		First, a mitosis detector is trained using labeled HPFs.
		Then, additional mitosis are mined from unannotated WSIs and finally, the mitosis detector is re-trained using patches from both labeled HPFs and unlabeled WSIs.
	}
	\label{fig:nets}
\end{figure*}

This network is trained using patches of 128x128, which have the label of either background or foreground (mitosis).
These patches are sampled from the following four sets depending on the training stage: \textit{BG-Rand}, \textit{BG-Hard}, \textit{FG-Lab} and \textit{FG-WSI} (see Fig.~\ref{fig:nets}a).

Set \textit{BG-Rand} contains patches, which are sampled randomly with uniform probability from HPFs such that their distance is greater than a fixed threshold from previously selected background patches and all mitotic cells in that HPF.

\textit{BG-Hard} consists of hard-negative mined patches, i.e. the false positive patches extracted by applying the model (@ 30k iteration) to training HPFs.
The detections for each HPF are post-processed using non-maximum suppression (NMS), with a distance of 50 pixels, to remove duplicate.

Set \textit{FG-Lab} contains patches centered at the annotated mitosis positions and set \textit{FG-WSI} includes mitosis patches extracted from the unlabeled whole-slide-images (WSIs), details of which are provided in Section \ref{method:mining}.

The network weights are optimized using ADAM optimizer \cite{Kingma2015}, which minimizes the cross-entropy loss.
We use a learning rate of 0.0001 for the first 20k iterations, which is lowered to 0.00001 for the next 30k iterations and for the final 10k iterations, it is dropped to 0.000001.
We use a batch size of 64 with a fixed ratio of mitotic:background patches (2:62).

For the initial model, which makes use of only the labeled data, we train it using patches from sets \textit{BG-Rand} and \textit{FG-Lab} for the first 30k iterations.
Then, hard-negative mining is used to obtain \textit{BG-Hard} patches from the training HPF. 
The next 30k iterations sample background patches from \textit{BG-Hard}.

For the final model, which utilizes both the labeled and unlabeled data, the foreground (mitosis) patches are sampled from \textit{FG-Lab} and \textit{FG-WSI} sets with a fixed ratio and the background patches are sampled from \textit{BG-Rand} for first 30k iterations, then hard-negative mining is used to obtain background patches \textit{BG-Hard} and the model is trained for another 30k iterations.

\vspace{-0.6em}
\subsubsection{Data augmentation:}
Contrast variation is one of the biggest challenges in automated analysis of histopathology images.
Slides prepared using different staining protocols and/or imaged using different scanners can have very different color distributions (see Fig. \ref{fig:datasets}).
We use contrast transfer \cite{Reinhard2001} to make our network robust to these variations.
During training, with a fixed probability, we change the mean and standard deviation (in LAB color-space) of the HPF from which a patch is sampled to the values from another randomly selected HPF.
We also use flipping, rotation and jitter to augment training samples.

\vspace{-0.6em}
\subsubsection{Testing:}
At test time, we apply our method in a fully-convolutional manner.
Since the test images can be too large to fit in the current GPU memory, we process them in windows of size 512x512.
We pad (64 pixels) images by mirroring and use overlapping (120 pixels) windows to handle zero-padding in convolutional layers.

Once we have the probability maps of all windows predicted by our network, we stitch and re-size them to the original image size.
This is followed by NMS with radius of 30 pixels to discard duplicate detections.
Finally, we remove all detections with score below 0.5.

\subsection{Mining mitosis from unlabeled WSIs}
\vspace{-0.6em}
\label{method:mining}
Current mitosis detectors typically have a high false negative rate at low values of false positive rate (see Fig.~\ref{pr-curve}), so we use unlabeled data only for mining mitosis samples.
In order to ensure that the training does not derail, it is important that the additional mitosis samples have very few false positives (FPs).
We achieve this by using test-time data augmentations (flipping and rotations), which removes many FPs with high score for only few of these transformations.

We apply our mitosis detector to whole-slide-images (WSIs) in a sliding window manner.
Processing WSIs can be computationally expensive and applying our mitosis detector on a single WSI can takes hours.
In order to reduce the computational demands, we skip windows which have mean RGB value greater than a fixed threshold.
Similarly, if the maximum score in a window is below a threshold for the first transformation, then additional transformations are skipped.

\vspace{-0.6em}
\section{Experiments}
\vspace{-0.6em}
\subsubsection{Dataset: }
\begin{figure}[!tb]
	\centering
	\subfloat[Center 1\label{ds-1}]{
		\includegraphics[width=0.19\textwidth]{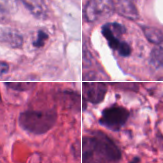}
	}\hspace{-0.7em}
	\subfloat[Center 2\label{ds-2}]{
		\includegraphics[width=0.19\textwidth]{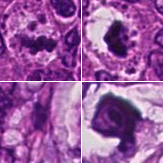}
	}\hspace{-0.7em}
	\subfloat[Center 3\label{ds-3}]{
		\includegraphics[width=0.19\textwidth]{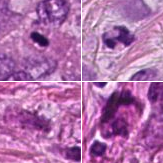}
	}\hspace{-0.7em}
	\subfloat[Aperio\label{ds-4}]{
		\includegraphics[width=0.19\textwidth]{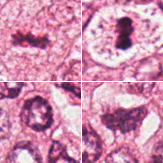}
	}\hspace{-0.7em}
	\subfloat[Hamamatsu\label{ds-5}]{
		\includegraphics[width=0.19\textwidth]{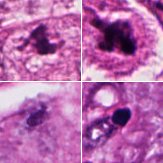}
	}
	\caption{{\footnotesize Top row: Mitotic samples. 
			Bottom row: Hard-negative samples. 
			a-c) are from TUPAC dataset and were collected from 3 different pathology centers.
			d-e) are from MITOS14 dataset and they were scanned using two different scanners.}}
	\label{fig:datasets}
\end{figure}

We evaluate our method on two public mitosis datasets: TUPAC \cite{tupac} and MITOS14 \cite{mitos}.
We split these datasets into 80\% train/20\% val sets at case level.
Fig. \ref{fig:datasets} shows some mitosis and hard-negative patches from these datasets.

TUPAC (\textbf{T}) training set contains 1,552 mitotic cells annotated in 656 HPFs, which are selected from 73 cases acquired from 3 different pathology labs.
This dataset also includes 500 WSIs from 500 cases in the training set which are not annotated for mitosis.
We use these WSIs as unlabeled data (\textbf{U}) for our experiments.
MITOS14 (\textbf{M}) training set consists of 1,502 mitotic cells marked in 1,200 HPFs from 11 cases, each scanned using two different scanners: Aperio Scanscope XT and Hamamatsu Nanozoomer 2.0-HT.

\vspace{-0.6em}
\subsubsection{Metrics: }
%\vspace{-0.6em}
We use the same criteria as TUPAC challenge \cite{tupac} for evaluating mitosis detection performance.
A detection is considered as true positive (TP) if it is within a radius of 30 pixels from any unmatched ground truth (GT).
All other detections are considered false positives (FP) and the GT annotations without any detection in 30 pixel radius are counted as false negative (FN).
Then, recall (R), precision (P) and F1-Score (F1) are computed.
When comparing different models, we use maximum F1-Score at any threshold value; this mitigates the problem of high sensitivity of F1-score to the selected threshold.

\begin{figure}[!tb]
	\centering
	\subfloat[Dataset size\label{datasize}]{
		\includegraphics[width=0.32\textwidth]{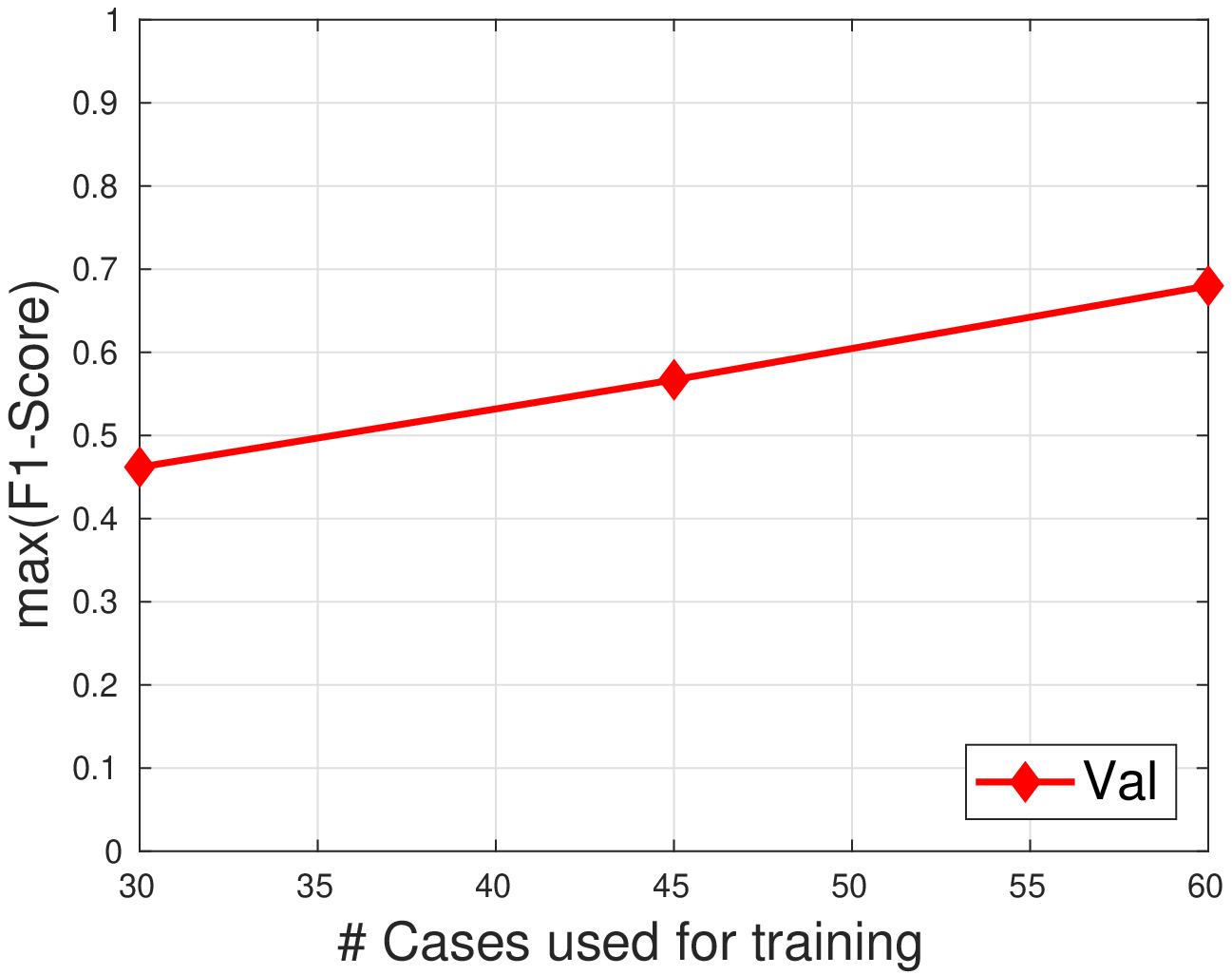}
	}\hspace{-1em}
	\subfloat[Validation error\label{val-error}]{
		\includegraphics[width=0.32\textwidth]{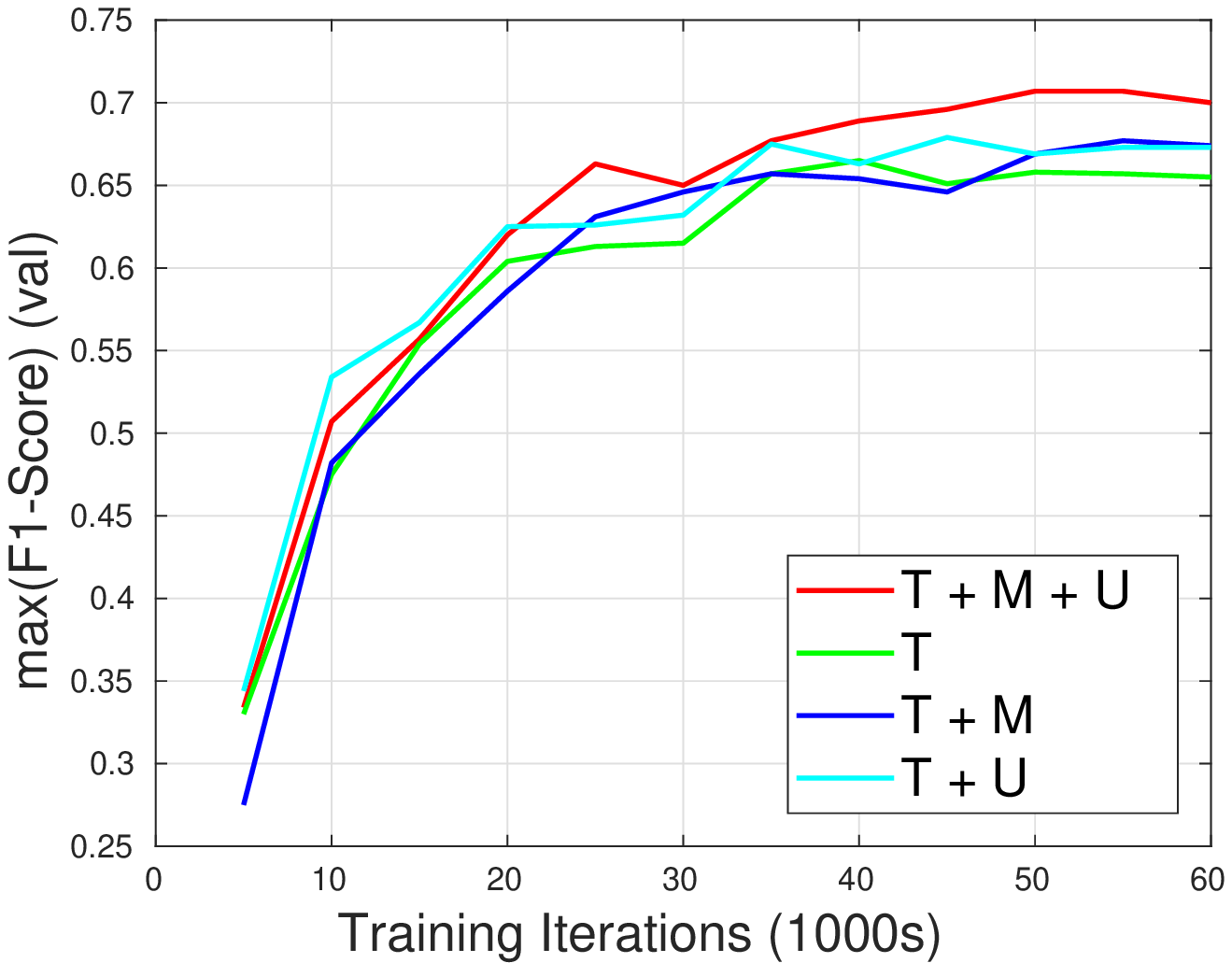}
	}\hspace{-1em}
	\subfloat[PR curve\label{pr-curve}]{
		\includegraphics[width=0.32\textwidth]{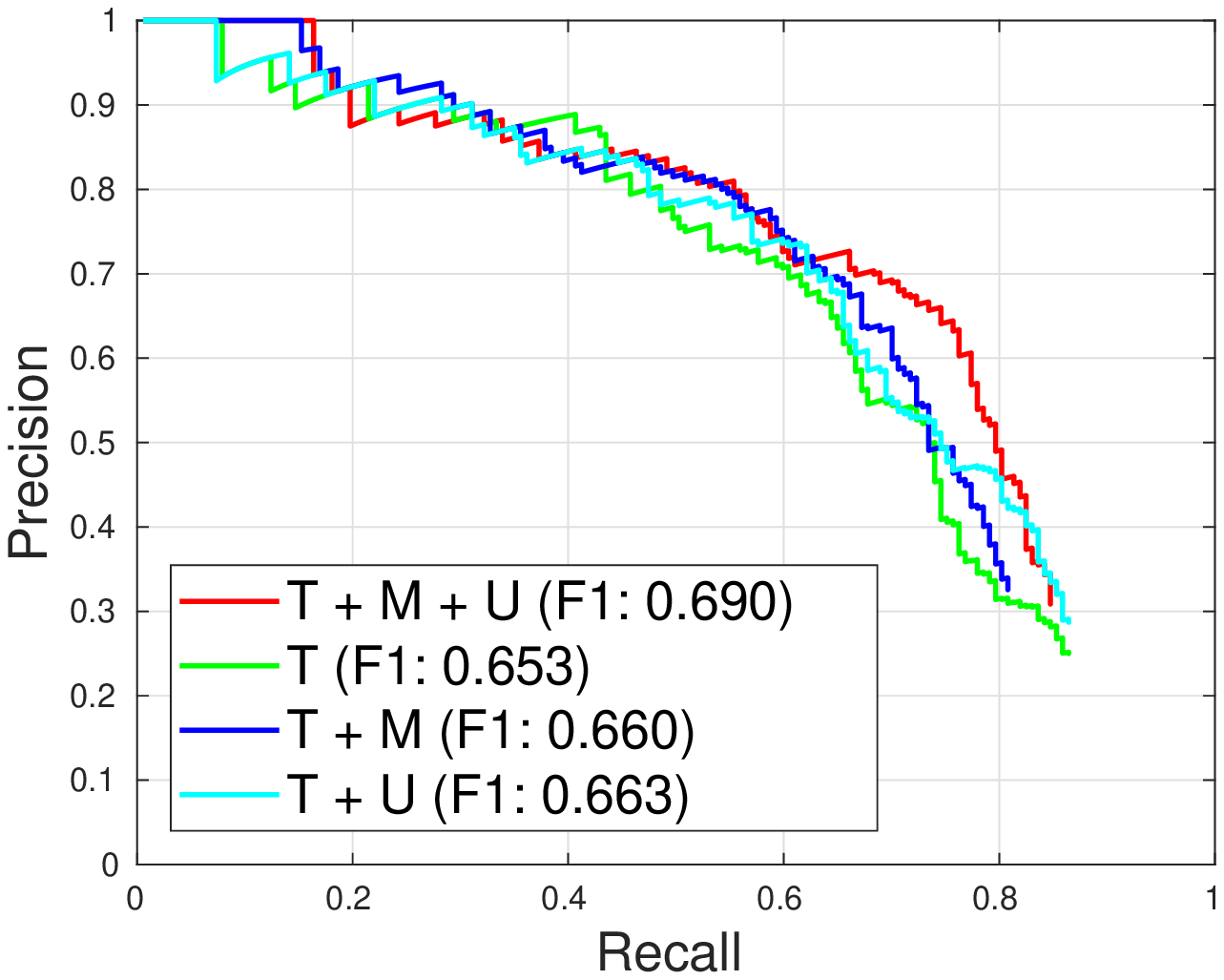}
	}
	\vspace{-0.1em}
	\caption{a) Performance improves almost linearly with the TUPAC training data size. b) TUPAC validation error during training for various models. c) Precision-Recall curve  (TUPAC-val). F1-Scores @ threshold=0.5 are listed in the legend.}
	\label{fig:results1}
	\vspace{-0.6em}
\end{figure}

\vspace{-0.6em}
\subsubsection{Dataset size:}
We select 13 cases as validation set and then train three models with the same hyper-parameters but with 30, 45 and 60 cases in the training set.
The results are plotted in Fig. \ref{datasize}, which shows that the performance improves almost linearly as we increase the training set size.

\vspace{-0.6em}
\subsubsection{Cross-dataset generalization:}
Table~\ref{tab:cross-data} presents performance on TUPAC (T) and MITOS14 (M) validation sets for models trained using various combinations of training sets.
The performance on TUPAC-val drops sharply when only MITOS14 and unlabeled data (\textbf{M} and \textbf{M+U} in Table~\ref{tab:cross-data}) is used for training, which can be explained by the fact that TUPAC data is collected from multiple pathology centers and as a result it has much larger variation than MITOS14.
The performance on MITOS14 is much more consistent irrespective of which training set is used for training, indicating that at least some cases in TUPAC data are representative of this dataset.

\setlength{\tabcolsep}{4pt}
%\begin{wraptable}{r}{5.5cm}
\begin{table*}[!tb]
\vspace{-0.5em}
	\begin{center}
		\caption{{\footnotesize Performance on both TUPAC/MITOS14-val sets for various combinations of training sets: TUPAC (T), MITOS14 (M) and unlabeled WSIs (U).}}
		\label{tab:cross-data}
		\begin{tabular}{ccc | ccc}
			Train set& T 		& M 		&Train set	& T 		& M\\
			& max(F1) 	& max(F1)	&			& max(F1) 	& max(F1)\\
			\hline
			T      	& 0.655 	& 0.552 	& T+U		& 0.673 	& 0.579\\ % T:+2.75%, M:4.89%
			M		& 0.378 	& 0.678 	& M+U		& 0.466 	& 0.674\\ % T:+23.28%, M:-0.59%
			T+M		& 0.653 	& 0.660 	&T+M+U 		& 0.684		& 0.655\\ % T:+4.75%, M:-0.75%
		\end{tabular}
	\end{center}
\end{table*}

\vspace{-0.6em}
\subsubsection{Impact of unlabeled WSIs:}
The experiments with unlabeled data (Table~\ref{tab:cross-data}) show that the performance on TUPAC-val set improves considerably for any combination of TUPAC and MITOS14 dataset, indicating that mitosis samples mined from unlabeled WSIs provide additional non-trivial information which is helpful for learning a better model.
For MITOS14-val on the other hand, the use of unlabeled data only helps when MITOS14-train data is not used during training.
This is due to the unlabeled data having much larger contrast variation, which makes the model generalize better (as performance on TUPAC always improves) but does not provide much additional help when training set already includes data from MITOS14.

Fig.~\ref{val-error} shows the performance on TUPAC-val during training for few models trained using different combinations of \textbf{T}, \textbf{M} and \textbf{U} sets and Fig.~\ref{pr-curve} shows the precision-recall curve for these models.
Hard-negative mining is performed at 30k iterations and it consistently improves the F1-score by 5-10\%.

%\end{wraptable} 
\vspace{-0.6em}
\subsubsection{Case-level performance:}
Mean and standard deviation of F1-Score at case level for models trained using only \textbf{T} (0.622$\pm$0.236), \textbf{T + M} (0.619$\pm$0.227) and \textbf{T + M + U} (0.636$\pm$0.191) also indicate that using only MITOS14 data in addition to TUPAC has little overall benefit as this data is from a different source.
However, when unlabeled data from TUPAC is used, it leads to much greater improvement in performance and the variation across cases reduces as well, indicating that this automatically-labeled data is better for training than the manually labeled data which comes from a different source.

\vspace{-0.6em}
\subsubsection{Comparison with state-of-the-art:}
Table~\ref{tab:comparison} compares the performance of our method with top ranked methods on both TUPAC and MITOS14 datasets.
On MITOS14, our method has much higher F1-score on validation set (0.642) compared to top ranked method on test set (0.482).
%However, these numbers are not directly comparable.
On TUPAC test set, our method ranks 3rd\footnote{Out of 20 methods listed at \url{http://tupac.tue-image.nl/node/62}
} with F1-score of 0.64.
We would like to point out that our submission consists of a single model (@threshold=0.5) with minimal post-processing (i.e. only NMS).
Techniques such as test-time augmentation and ensembling can potentially improve performance considerably.
With ensembling alone leading to $\sim$10\% improvement in F1-score of some methods \cite{Chen2016,Tellez2018}) on this task.

\begin{table*}[!tb]
	\vspace{-2em}
	\begin{center}
		\caption{{\footnotesize Performance on TUPAC/MITOS14. The results in first two rows are on validation set, while the rest are on test sets.}}
		\label{tab:comparison}
		\begin{adjustbox}{max width=\textwidth}
		\begin{tabular}{cccc | cccc}
			 	& \multicolumn{3}{c|}{TUPAC} 		&		& \multicolumn{3}{c}{MITOS14} \\
								& F1 	& Recall 	& Precision	& 				& F1 		& Recall 	& Precision\\
		\hline
		Ours (T+M+U) 			& 0.690 & 0.661&0.722 	& Ours (T+M+U)					& 0.642 	& 0.605	& 0.683		\\
		Ours (T) 				& 0.653 & 0.621 	&0.688 		& Ours (M)						& 0.620 	& 0.496	& 0.828		\\
		\hline
		Lunit \cite{Paeng2017}  & 0.652 & 	-		&		- 	& CasNN	\cite{Chen2016}			& 0.482 	& 0.507	& 0.460	\\
		IBM \cite{Zerhouni2017} & 0.648 & 	-		&		- 	& DeepMitosis \cite{Li2018}		& 0.437 	& 0.443	& 	0.431	\\
		Ours (T+M+U) 			& 0.640 & 0.671		&0.613 	& -					&  -	& -	& 	-	\\		
		\end{tabular}
		\end{adjustbox}
	\end{center}
\end{table*}

\vspace{-0.9em}
\section{Discussion}
\vspace{-0.4em}
In this paper, we have presented a semi-supervised mitosis detection method, which benefits from a large amount of unlabeled data, that is readily available in histopathology.
This method has the potential to benefit other detection and segmentation tasks in many medical and biological research areas, where labeled data is scarce and unlabeled data is abundant.
It is clear from current methods \cite{Khoreva2017,Radosavovic2017} and our own results that once a method reaches a sufficient performance level, it can be used to generate pseudo-labels for unlabeled data, which can then be used to re-train it, improving performance.
However, it is worth investigating, where that sufficient performance level lies for various tasks.

An important future direction is to explore the upper limit of performance that is attainable with current annotation procedures.
This upper-limit exists, due to the unavoidable noise in pathologist mitosis annotation.
State-of-the-art methods already obtain similar performance to that of pathologists, but these methods have poor generalization when applied to data from different sources \cite{Tellez2018}.
Using larger and more diverse datasets can resolve this issue and semi-supervised techniques can be highly beneficial in these scenarios.
In addition, more objective annotation techniques (e.g. using PHH3 labeling \cite{Veta2015}) have the potential of improving performance considerably, which may result in future automated methods outperforming experienced pathologists.

\vspace{-0.6em}
\bibliographystyle{splncs03saad}
\bibliography{compay18}
\end{document}